\documentclass[conf]{new-aiaa}
\usepackage[utf8]{inputenc}

\usepackage{graphicx}
\usepackage{amsmath}
\usepackage[version=4]{mhchem}
\usepackage{siunitx}
\usepackage{longtable,tabularx}
\usepackage{booktabs}
\usepackage{float}
\usepackage{placeins}
\usepackage{dsfont}
\usepackage{subfigure}

\usepackage{comment}
\usepackage{graphicx}
\usepackage{amsmath}
\usepackage[version=4]{mhchem}
\usepackage{siunitx}
\usepackage{longtable,tabularx}
\usepackage{xcolor}
\usepackage{cleveref}
\usepackage[ruled]{algorithm2e}% 
\makeatletter
\renewcommand{\@algocf@capt@plain}{above}
\usepackage{algpseudocode}
\usepackage{soul}
\usepackage{makecell}
\usepackage{multirow}
\DeclareMathOperator{\E}{\mathbb{E}}

\definecolor{rev1}{HTML}{FF999A} 
\definecolor{rev2}{HTML}{F3F298} 
\definecolor{rev3}{HTML}{B2E0AE}
\definecolor{other}{HTML}{C8C7FF} 
\sethlcolor{rev1}

\soulregister\cite7
\soulregister\citep7
\soulregister\citet7
\soulregister\ref7
\soulregister\eqref7

\title{Quantum Computing-Based Wasserstein Generative Adversarial Networks for Airfoil Parameterization}

\author[]{Abdur Rahman Ovy~\footnote{Doctoral Student, Department of Mechanical and Aerospace Engineering, AIAA Student Member}}
\author[]{Xiaosong Du~\footnote{Assistant Professor, Department of Mechanical and Aerospace Engineering, AIAA Member.}}
\affil[]{Missouri University of Science and Technology, Rolla, MO 65409}

\begin{document}

\maketitle

\begin{abstract}

Airfoil geometry parameterization is essential in aerodynamic design optimization.
Conventional parameterization methods formulate a large design space to guarantee real optimal airfoil designs are included.
As a result, optimization search in the large space suffers from low efficiency and modest feasible region, which could further incur failures to convergence. 
Generative adversarial networks (GAN), aiming to learn the statistical distribution of existing data, pave a new avenue for shape generation and airfoil parameterization.
GAN manages to implicitly reduce design space by generating only smooth, realistic airfoils. 
Meanwhile, quantum computing has been emergingly exhibiting great potential for achieving superior performance in specific types of engineering problems but has not been introduced to airfoil parameterization.
To fill this gap, this work develops quantum computing-based Wasserstein GAN (Q-WGAN) and investigate the performance for airfoil parameterization. 
The Q-WGAN adopts parameterized quantum circuits as the generator to promote the generative performance, while the discriminator remains classical.
The developed Q-WGAN is demonstrated on a dataset of optimal airfoils parameterized by free-form deformation (FFD) control points, resulting from 992 airfoil aerodynamic design optimizations. 
Results show that Q-WGAN achieves 92.1\% mean fitting accuracy on the optimal FFD control points and 99.2\% accuracy on airfoil coordinates, comparable with the classical WGAN performance of 92.28\% on FFD control points and 99.26\% on airfoil coordinates.
Results also reveal that Q-WGAN and classical WGAN learn significantly different representations of the FFD control points.
The Q-WGAN discovers a more compact distribution, while the classical WGAN covers a broader area but focuses more on the bounds of each FFD control point.
Due to this difference, Q-WGAN achieves lower Jensen-Shannon divergence, Fréchet distance, and Wasserstein distance than the classical WGAN, indicating an improved agreement with the reference airfoil distribution.
Therefore, this work presents a convincing evidence that Q-WGAN makes a viable option for parameterizing airfoils and possesses a great potential for other relevant engineering applications.
\end{abstract}

\section{Introduction}

Airfoil parameterization, a key component of aerodynamic shape optimization, plays a critical role in improving aircraft performance.
Conventional airfoil parameterization methods, such as B-spline curves, free-form deformation (FFD), and PARSEC method, have exhibited success in airfoil shape design~\cite{li2022machine,chen2019aerodynamic,chen2021deep}. 
However, naively increasing the number and expanding the bounds of design variables to ensure geometric flexibility could incur extremely inefficient optimization, large portion of infeasible region, or even failed optimization convergence. 

To tackle the above challenge, dimensional reduction has therefore become a viable strategy for constructing more compact aerodynamic design spaces.
Traditional dimensionality reductions methods, such as principal component analysis, singular value decomposition, proper orthogonal decomposition, and active-subspace-type techniques have been investigated extensively, while more recent studies have increasingly adopted generative artificial intelligence~\cite{yasong2018global, viswanath2011dimension, alswaitti2022dimensionality, jun2020application, cinquegrana2018investigation, tejero2023aerodynamic, wu2024high, bo2024data}.
\citet{chen2020airfoil} introduced B{\'e}zier-GAN for compact airfoil parameterization. 
\citet{du2020b} developed B-spline-based GAN and reported less than 1\% fitting error towards the UIUC airfoil database.
\citet{wang2021airfoil} proposed Airfoil GAN for latent-space airfoil reconstruction and optimization. 
\citet{chen2021deep} developed FFD-GAN for compact generative parameterization of aerodynamic geometries.
These work leveraged the feature of implicit dimensionality reduction of GAN, \emph{i.e.}, generating only realistic geometries and automatically removing unrealistic ones from the design space. 
\citet{sisk2026physics} further demonstrated the two-fold dimensionality reduction, \emph{i.e.}, both implicit and explicit reductions, in aerospace optimization, reducing a 41-variable takeoff-trajectory problem to three variables while retaining high fitting accuracy and feasible-space coverage.

Although GAN-based parameterization provide considerable flexibility, conventional GAN training can suffer from unstable convergence, vanishing gradients, and mode collapse~\cite{goodfellow2014generative, wu2019pwgan, chen2020improvement}.
Wasserstein GAN (WGAN) was introduced by \citet{arjovsky2017wasserstein} to improve adversarial training by replacing the conventional divergence-based objective with the Wasserstein, or Earth-Mover, distance, while \citet{gulrajani2017improved} later proposed WGAN-GP using a gradient penalty to improve enforcement of the Lipschitz constraint.
Wasserstein-based models have also been applied to airfoil generation.
\citet{yonekura2024airfoil} used conditional WGAN-GP to generate smooth airfoils satisfying prescribed aerodynamic requirements, and related studies have incorporated Wasserstein objectives into conditional and hybrid generative models for aerodynamic inverse design.

Meanwhile, advances in quantum computing-based machine learning have motivated quantum extensions of classical generative models.
\citet{lloyd2018quantum} introduced the theoretical framework for quantum generative adversarial networks (QGAN), while \citet{situ2020quantum} and \citet{zoufal2019quantum} demonstrated hybrid architectures in which parameterized quantum circuits acted as generators and classical neural networks performed discrimination.
\citet{huang2021experimental} experimentally demonstrated QGAN-based image generation and compared quantum and classical generative models.
\citet{chakrabarti2019quantum} extended Wasserstein adversarial learning to the quantum setting and reported improved robustness and scalability relative to earlier QGAN formulations in their quantum-data experiments.
Reviews of QGAN research show that most existing quantum generative studies focus on quantum states, generic probability distributions, or image-generation benchmarks rather than aerodynamic shape representation~\cite{kushwaha2020study, ngo2023survey}.

To fill the gap above, this work investigates the performance of quantum computing-based WGAN (Q-WGAN) on airfol parameterization and compares performance with classical WGAN.
The Q-WGAN constructs the generator networks using quantum circuits while keeping the discriminator networks classical to save quantum qubits (see Sec.~\ref{sec:qwgan}).
The performance metrics include FFD and coordinate fitting accuracy, Jensen–Shannon divergence (JSD)~\cite{lin1991divergence}, Kullback–Leibler (KL) divergence~\cite{kullback1951information}, Fréchet distance (FD)~\cite{heusel2017gans}, Wasserstein distance, and t-distributed stochastic neighbor embedding (t-SNE)~\cite{van2008visualizing}.
Please refer to Sec.~\ref{sec:results} for more details.

The rest of this paper is organized as follows. 
Section~\ref{sec:method} describes the classical WGAN and Q-WGAN, followed by performance metrics. 
Section~\ref{sec:results} demonstrates and compares the performance of Q-WGAN on airfoil parameterization against classical WGAN. 
This paper ends with conclusions in Sec.~\ref{sec:conclusion}.

\section{Methodology}
\label{sec:method}

This section elaborates on the classical WGAN and Q-WGAN.
This section also introduces the performance metrics, including fitting optimization accuracy and distribution metrics. 

\subsection{Classical Wasserstein Generative Adversarial Network}

\begin{figure}[!htbp]
\centering
\subfigure[WGAN workflow]{%
    \includegraphics[width=0.45\linewidth]{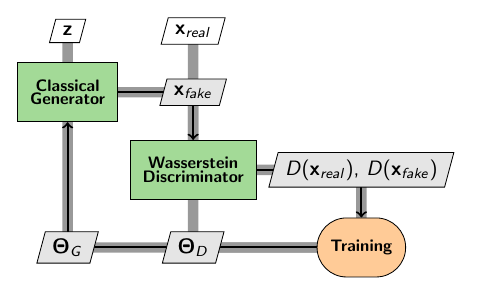}%
    \label{fig:Classical WGAN_XDSM}%
}\hfill
\subfigure[Classical generator architecture]{%
    \includegraphics[width=0.55\linewidth]{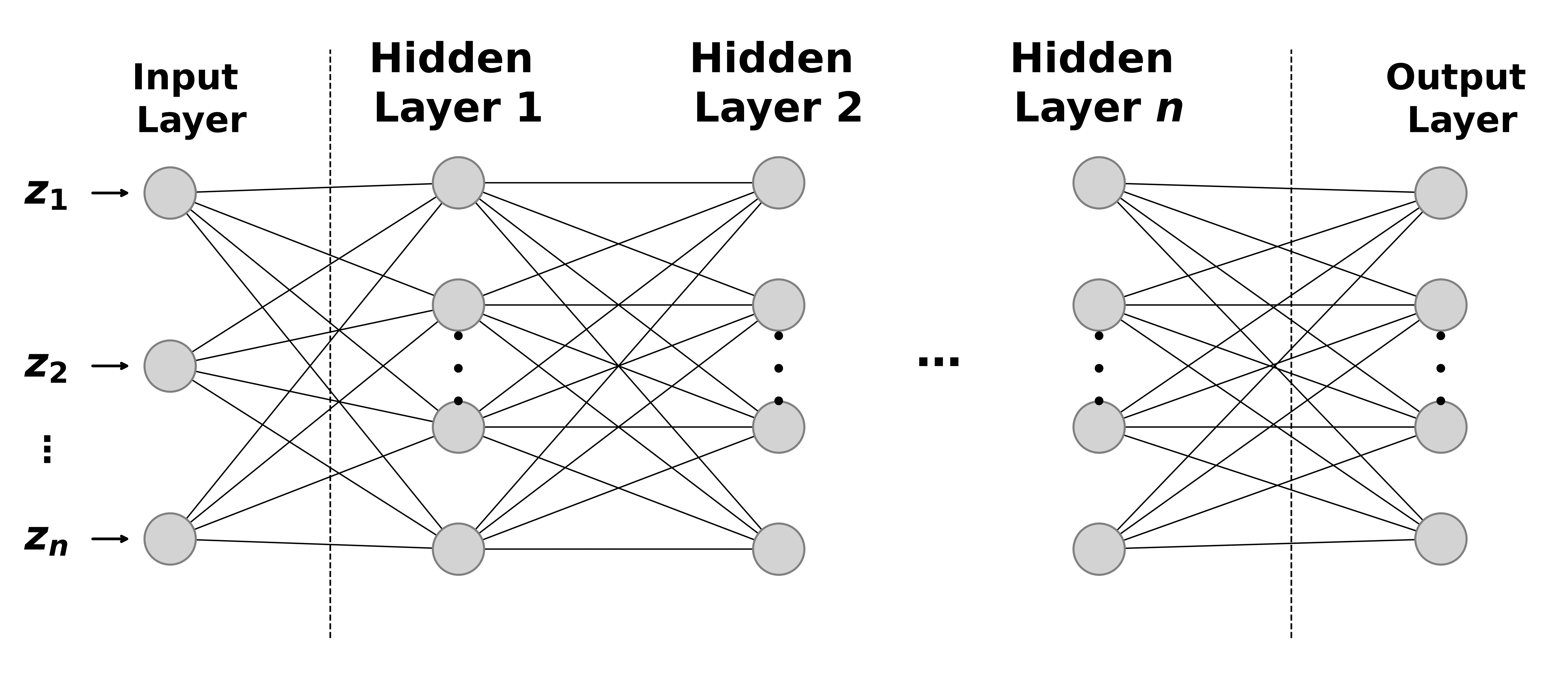}%
    \label{fig:cgen}%
}
\caption{Workflow and generator architecture of the Classical Wasserstein GAN.}
\label{fig:cwgan}
\end{figure}

GAN (Fig.~\ref{fig:Classical WGAN_XDSM}) consists of two competing neural networks: a generator that maps samples from a latent space to synthetic data; and a discriminator that attempts to distinguish generator-generated data from real samples.
During adversarial training, both networks are updated iteratively until the generated distribution approaches the reference data distribution~\cite{goodfellow2014generative}.
Conventional GAN training, however, can suffer from unstable gradients and mode collapse.
WGAN addresses these issues by replacing the discriminator with a critic that estimates the Wasserstein-1 distance between the real and generated distributions.

Specifically, the generator is trained by minimizing the Wasserstein objective:
\begin{equation}
\label{eq:generator_loss}
L_G = -\E_{\mathbf{z}\sim p_z}
\left[D\left(G(\mathbf{z})\right)\right],
\end{equation}
where $L_G$ is generator loss, $\E_{\mathbf{z}\sim p_z}$ refers to expectation over latent vectors \(z\) sampled from the latent distribution \(p_z\), $D$ is Wasserstein discriminator function and $G$ is generator function.

The discriminator receives both real and generated data and assigns a scalar score to each sample.
During training, it learns to assign higher scores to real samples than to generated samples.
Unlike the discriminator of a conventional GAN, no sigmoid activation is applied to the wasserstein discriminator output.
The wasserstein discriminator updates five times per generator update.
The discriminator is trained by minimizing the following Wasserstein objective:
\begin{equation}
\label{eq:critic_loss}
L_D =
\E_{\mathbf{z}\sim p_z}\left[D\left(G(\mathbf{z})\right)\right]
-
\E_{\mathbf{x}\sim p_{\text{data}}}\left[D(\mathbf{x})\right],
\end{equation}
where $L_D$ is discriminator loss, $\E_{\mathbf{x}\sim p_{\text{data}}}$ refers to expectation over real samples $\mathbf{x}$ drawn from the real-data distribution \(p_{data}\).

In this work, the classical WGAN is used to generate
18 FFD control points, which parameterize each airfoil.
The classical generator is a fully connected neural network with layer dimensions $(3 \to 256 \to 512 \to 1024 \to 18)$.
Leaky-ReLU activation functions are used in the hidden layers, together with batch normalization, while a hyperbolic tangent activation is applied to the output layer to match the normalized FFD-variable range (Fig.~\ref{fig:cgen}).

The original weight-clipping formulation of WGAN is adopted, with the discriminator parameters constrained to ([-0.01,0.01]) after each discriminator update.
The discriminator is updated five times for each generator update, and both the generator and discriminator are trained using the Adam optimizer.
Training is done with different combinations of learning rates and got the best result with a learning rate of $2 \times 10^{-4}$ for both generator and discriminator.
The training dataset consists of 20 optimal airfoils from 20 airfoil aerodynamic design optimizations (Sec.~\ref{sec:results}). 
The details of the classical WGAN are presented in Table~\ref{tab:Classical WGAN_parameters}
\begin{table}[!htbp]
\centering
\caption{Classical WGAN architecture and training parameters}
\label{tab:Classical WGAN_parameters}
\begin{tabular}{lcc}
\toprule
Parameters & Generator & Discriminator \\
\midrule
Input  & $\mathbf{z} \sim \mathcal{U}([0,1]^3)$ & 18-dimensional FFD vector \\
Architecture & $[3 \to 256 \to 512 \to 1024 \to 18]$ & $[18 \to 1024 \to 512 \to 256 \to 1]$ \\
Trainable parameters & 679,442 & 675,841 \\
Learning rate & $2 \times 10^{-4}$ & $2 \times 10^{-4}$ \\
Hidden activation & LeakyReLU $(\alpha = 0.2)$ & LeakyReLU $(\alpha = 0.2)$ \\
Normalization & Batch normalization & N/A \\
Output & 18-dimensional FFD vector & Scalar score \\
Output activation & Tanh & Linear \\
\bottomrule
\end{tabular}
\end{table}

\subsection{Quantum Computing-Based Wasserstein Generative Adversarial Network}
\label{sec:qwgan}

\begin{figure}[!htbp]
\centering
\subfigure[Q-WGAN workflow]{%
    \includegraphics[width=0.48\linewidth]{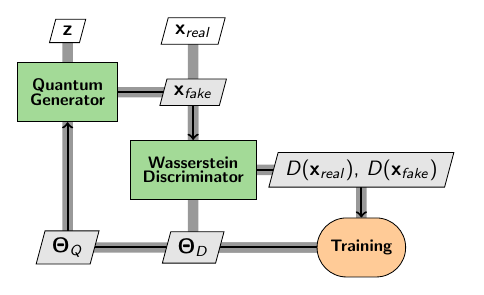}%
    \label{fig:QWGAN_XDSM}%
}\hfill
\subfigure[Quantum generator architecture]{%
    \includegraphics[width=0.48\linewidth]{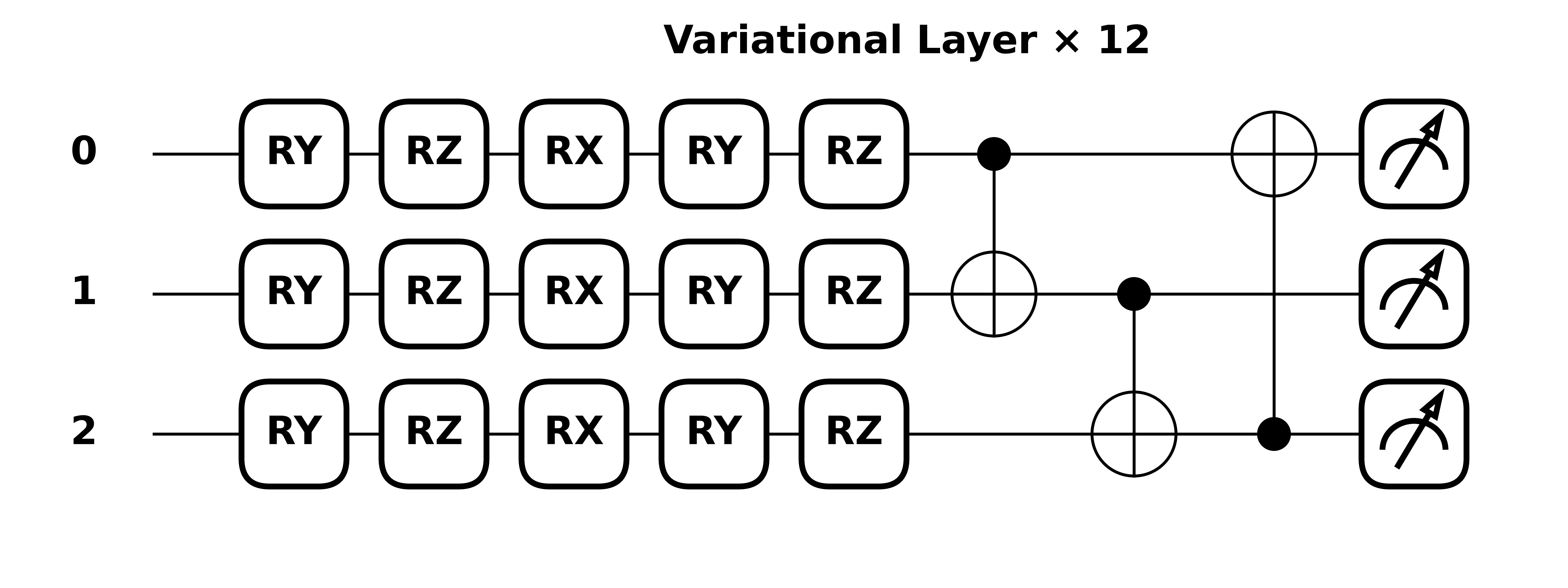}%
    \label{fig:qgen}%
}
\caption{Workflow and generator architecture of the Q-WGAN.}
\label{fig:qwgan}
\end{figure}

Q-WGAN extends the adversarial learning framework by replacing one or both classical neural networks with parameterized quantum circuits (PQC).
In this work, the classical generator of the Classical WGAN is replaced by PQC while retaining a classical Wasserstein discriminator (Fig.~\ref{fig:QWGAN_XDSM}), resulting in a hybrid Q-WGAN.
A three-dimensional latent vector is supplied to a three-qubit quantum generator:
\begin{equation}
\mathbf{z} = [z_1, z_2, z_3]^{\mathsf{T}},
\qquad
z_i \sim \mathcal{U}(0,1),
\quad i = 1,2,3.
\end{equation}
Each latent variable is encoded on the corresponding qubit using successive $R_Y$  and  $R_Z$ rotations,where, $R_Y$ rotates qubit $i$ by an angle of $\pi z_i$ and $R_Z$ rotates by an angle of $2\pi z_i$.
The encoding transforms the classical latent variables into the quantum state before application of the trainable variational circuit.

The variational portion of the generator consists of repeated quantum layers.
Each layer applies trainable $R_X$, $R_Y$, and $R_Z$ rotations to every qubit, followed by a ring of controlled-NOT gates to introduce entanglement between the qubits.
For three qubits, the entangling pattern is $q0 \to q1,   q1 \to q2,   q2 \to q0$.
In each connection, the first qubit acts as the control qubit and the second as the target qubit.
This closed-loop structure allows information encoded in one qubit to influence the states of the other qubits, enabling the circuit to represent correlations among the latent variables.
The combination of trainable single-qubit rotations and ring entanglement is repeated across the variational layers of the quantum generator.

Therefore, each variational layer contains nine trainable quantum parameters, corresponding to three rotation parameters for each of the three qubits.
A 12-layer circuit is used for the final Q-WGAN comparison (Fig.~\ref{fig:qgen}). 
Consequently, the number of trainable quantum parameters of final quantum circuit is
\begin{equation}
N_Q = 12 \times 3 \times 3 = 108.
\end{equation}

After the variational circuit, the expectation value of the Pauli-(Z) operator is measured on each qubit using Eq.~\ref{eq:exp}, producing the three-dimensional quantum output vector using Eq.~\ref{eq:vector}.
Because the desired airfoil representation contains 18 FFD variables, the three measured expectation values are subsequently passed through a classical fully connected layer from Eq.~\ref{eq:ffd_output}, where $\mathbf{W}\in\mathbb{R}^{18\times 3}$ and $\mathbf{b}\in\mathbb{R}^{18}$.
\begin{equation}
\label{eq:exp}
m_i = \left\langle Z_i \right\rangle,
\qquad i = 1,2,3.
\end{equation}
\begin{equation}
\label{eq:vector}
\mathbf{m} = [m_1, m_2, m_3]^{\mathsf{T}},
\end{equation}
\begin{equation}
\label{eq:ffd_output}
\hat{\mathbf{x}}_{\mathrm{FFD}}
= \tanh\left(\mathbf{W}\mathbf{m} + \mathbf{b}\right),
\end{equation}
where $\hat{\mathbf{x}}_{\mathrm{FFD}}$ is generated or reconstructed 18-dimensional FFD vector, $\mathbf{W}$ means classical output-layer weight matrix, $m$ refers to quantum measurement output vector.
The hyperbolic tangent activation maps the generated values to the normalized FFD-variable range.
The classical output layer contributes $(3\times18+18=72)$ trainable parameters.
Thus, the 12-layer Q-WGAN generator contains a total of (108+72=180) trainable parameters.

The generated 18-dimensional FFD vectors and the real FFD samples are evaluated by a classical Wasserstein discriminator.
As in the classical WGAN, no sigmoid activation is applied to the discriminator output because the discriminator produces an unrestricted scalar score rather than a probability.
The Q-WGAN is trained using the same Wasserstein objectives,
\begin{align}
L_D &=
\E_{\mathbf{x}\sim p_{\text{data}}}
\left[D(\mathbf{x})\right]
-
\E_{\mathbf{z}\sim p_z}
\left[D\left(G_{\mathrm{Q}}(\mathbf{z})\right)\right],
\label{eq:qwgan_critic_loss}
\\
L_G &=
-\E_{\mathbf{z}\sim p_z}
\left[D\left(G_{\mathrm{Q}}(\mathbf{z})\right)\right],
\label{eq:qwgan_generator_loss}
\end{align}
where $G_Q$ denotes the hybrid quantum generator.
The original weight-clipping formulation of WGAN is used, with the discriminator weights restricted to $[-0.01,0.01]$.
The Q-WGAN is trained with different combinations of learning rates and got the best result with a generator learning rate of $(7.5\times10^{-4})$ and a discriminator learning rate of $(1\times10^{-4})$.
Both networks are optimized using the Adam optimizer~\cite{kingma2014adam}.

The quantum circuit is implemented using \texttt{PennyLane} with the \texttt{default.qubit} simulator.
Analytic expectation values are used by setting the number of measurement shots to None.
Therefore, the expectation values are evaluated directly from the simulated quantum state rather than estimated using finite-shot sampling.
The architecture and training parameters of the Q-WGAN are summarized in Table~\ref{tab:qwgan_parameters}.
\begin{table}[H]
\centering
\caption{Q-WGAN architecture and training parameters.}
\label{tab:qwgan_parameters}
\small
\setlength{\tabcolsep}{3pt}
\begin{tabularx}{\linewidth}{|>{\raggedright\arraybackslash}X|l|>{\raggedright\arraybackslash}X|l|}
\hline
\multicolumn{2}{|c|}{\textbf{Generator}} & \multicolumn{2}{c|}{\textbf{Discriminator}} \\ \hline
Input & $\mathbf{z} \sim \mathcal{U}([0,1]^3)$ & Input & 18-dimensional FFD vector \\ \hline
Input encoding & $R_Y(\pi z_i)$ and $R_Z(2\pi z_i)$ & Architecture & $[18 \to 64 \to 32  \to 1]$ \\ \hline
Entanglement & CNOT & Hidden activation & LeakyReLU $(\alpha = 0.2)$ \\ \hline
Variational gates & $R_Y$ , $R_X$ , $R_Z$ & Discriminator updates per generator update & 5 \\ \hline
Circuit depth & 12 & Optimizer & Adam \\ \hline
Quantum measurement & $\left\langle Z_1 \right\rangle, \left\langle Z_2 \right\rangle, \left\langle Z_3 \right\rangle$ & Batch size & 4 \\ \hline
Classical output layer & $3 \to 18$ & Loss function & Wasserstein loss \\ \hline
Output activation & Tanh & Output activation & Linear \\ \hline
Classical-head parameters & 72 & Weight-clipping range & [-0.01,0.01] \\ \hline
Trainable parameters & 180 & Trainable parameters & 3329 \\ \hline
Output & 18-dimensional FFD vector & Output & Scalar score \\ \hline
Learning rate & $7.5 \times 10^{-4}$ & Learning rate & $1 \times 10^{-4}$ \\ \hline
\end{tabularx}
\end{table}

\subsection{Latent-Space Optimization and Reconstruction Accuracy}
\label{sec:II.D}
To test whether a trained generator maintains sufficient variability compared with the training data, this work conducts fitting optimizations to reconstruct real airfoil data.
Differential evolution (DE)~\cite{storn1997differential}, a population-based global optimization algorithm, is employed to search the bounded latent space (\emph{i.e.}, [$\mathbf{0}, \mathbf{1}$] in this work).
DE does not require derivatives with respect to the latent variables and can explore multiple regions of the latent space simultaneously.

The latent-search objective is defined using the relative squared error,
\begin{equation}
\mathcal{L}_{2}(\mathbf{z}) =
\frac{\sum_{i=1}^{18} \left({x}_i - \hat{{x}}_i\right)^2}
{\sum_{i=1}^{18} {x}_i^2},
\end{equation}
where ${x}_i$ is the $i$th real FFD control point of a target airfoil, $\hat{x}_i$ is the $i$th generated FFD control point.
DE minimizes $\mathcal{L}_2$ with respect to the three latent variables.
The optimization bounds are ([0,1]) for each latent variable, with a population size of 150, a maximum of 10,000 iterations, a convergence tolerance of $(10^{-4})$, and final solution polishing enabled. 
Because DE is population based, a single DE optimization is performed for each target airfoil rather than using multiple independent latent-space restarts.

The final reconstruction performance is reported using relative $\ell_1$ accuracy for FFD control points and reconstructed airfoil coordinates:
\begin{equation}
Acc = 100\% -
\frac{1}{N}\sum_{i=1}^N\frac{\left|\mathbf{x}_i - \hat{\mathbf{x}}_i \right|_1}{\left|\mathbf{x}_i \right|_1},
\end{equation}
where $|\cdot|_1$ is the $\ell_1$ norm operator. 
This two-stage evaluation distinguishes between reconstruction quality in the reduced FFD design space and reconstruction quality of the resulting physical airfoil geometry.
The same DE procedure and accuracy definitions are applied to both the classical WGAN and Q-WGAN, enabling a fair comparison between the two types of models.

\subsection{Distribution Evalution Metrics}
In addition to reconstruction accuracy, the quality of the learned generative distributions is evaluated to assess the generative performance.
The resulting generated distributions are compared with the real data using JSD, KL divergence, Wasserstein distance, FD, and t-SNE.
Lower values of JSD, Wasserstein distance, FD, and KL divergence indicate greater agreement between the generated and reference distributions.

The Jensen–Shannon divergence provides a symmetric measure of similarity between two probability distributions.
For a reference distribution (P) and generated distribution (Q), the intermediate distribution is defined as $M = \frac{P+Q}{2}$.
and the Jensen–Shannon divergence is calculated as
\begin{equation}
\operatorname{JSD}(P,Q)
= \frac{1}{2}\operatorname{KL}(P \,\|\, M)
+ \frac{1}{2}\operatorname{KL}(Q \,\|\, M),
\end{equation}
where KL denotes the Kullback–Leibler divergence.
In this study, the JSD is evaluated independently for each of the FFD control points through histogram-based probability distributions, and the mean JSD value across all FFD variables is reported as the overall JSD.

Kullback–Leibler divergence quantifies the directional difference between the reference and generated probability distributions.
It is defined as
\begin{equation}
\operatorname{KL}(P \,\|\, Q)
= \sum_{i=1}^{N} P(i)
\ln\left(\frac{P(i)}{Q(i)}\right).
\end{equation}
The index \(i\) denotes the discrete histogram bin used to estimate the probability distributions
Because KL divergence is asymmetric, both directions are evaluated:
\[
\operatorname{KL}(\text{reference} \,\|\, \text{generated})
\]
and,
\[
\operatorname{KL}(\text{generated} \,\|\, \text{reference})
\]
The first direction emphasizes regions of the reference distribution that are insufficiently represented by the generator, whereas the reverse direction is more sensitive to generated probability mass occurring in regions with limited reference support.

Wasserstein distance represents the minimum amount of probability mass that must be transported, multiplied by the distance over which it is moved, to transform the generated distribution into the reference distribution.
For the cumulative distribution functions $(F_P)$ and $(F_Q)$, the first-order Wasserstein distance can be expressed as
\begin{equation}
W_1(P,Q)
= \int_{-\infty}^{\infty}
\left|F_P(x) - F_Q(x)\right|\,\mathrm{d}x.
\end{equation}
The one-dimensional Wasserstein distance is calculated independently for each FFD control point, 
and the mean Wasserstein distance is used provide an overall measure of distributional discrepancy.

Fréchet distance is used to compare the overall statistical structure of the generated and reference samples by considering their means and covariance matrices.
Assuming the reference and generated FFD samples are characterized by $((\mu_r, \sigma_r))$ and $((\mu_g, \sigma_g))$, respectively, the Fréchet distance is calculated as
\begin{equation}
\mathrm{FD}
=
\left\|\boldsymbol{\mu}_r - \boldsymbol{\mu}_g\right\|_2^2
+
\operatorname{Tr}\left[
\boldsymbol{\sigma}_r + \boldsymbol{\sigma}_g
- 2\left(
\boldsymbol{\sigma}_r
\boldsymbol{\sigma}_g
\right)^{1/2}
\right],
\end{equation}
where $\mu$ and $\sigma$ represent the corresponding mean vectors and covariance matrices.
Unlike the individual-feature JSD and Wasserstein calculations, the FD incorporates relationships among the FFD variables through the covariance matrices.

Moreover, t-SNE is employed to visualize high-dimensional joint structure in lower dimensions.
Unlike the preceding scalar metrics, t-SNE is used only as a visualization tool and is not interpreted as a quantitative distance measure.
Before applying t-SNE, each FFD variable is standardized using statistics obtained from the baseline dataset.
The same transformation is then applied to the Q-WGAN and classical WGAN samples.
The standardized reference and generated samples are subsequently embedded into a common two-dimensional space.
This visualization enables qualitative assessment of whether generated samples occupy neighborhoods similar to those represented by the baseline airfoils and whether the two generative models learn different regions or structures of the original design space.

\section{Results and Discussion}
\label{sec:results}
\subsection{Airfoil Dataset and Reference Problem Formulation}

The airfoil dataset used in this study is generated from the aerodynamic design problem summarized in Table~\ref{tab:optimization}.
The reference formulation considers airfoil shape optimization over a Mach-number range of [0.6, 0.7] and a Reynolds-number range of $[1.0\times10^6, 6.5\times10^6]$, with an allowable FFD-point variation of $\pm0.05$.
The coefficient of drag is minimized while aerodynamic and geometric constraints are imposed on lift coefficient, pitching-moment coefficient, airfoil volume, thickness.
The reference problem is formulated using the \texttt{MACH-Aero}~\cite{martins2022aerodynamic} aerodynamic shape optimization framework, which employs FFD through pyGeo~\cite{hajdik2023pygeo} for geometric parameterization and uses ADflow~\cite{mader2020adflow} for aerodynamic analysis within a gradient-based optimization framework.
\begin{table*}[!htbp]
\centering
\caption{Optimization variables, objectives, and constraints}
\label{tab:optimization}
\small
\begin{tabularx}{\textwidth}{@{}l c X c@{}}
\toprule
& \textbf{Variable} & \textbf{Description} & \textbf{Quantity} \\
\midrule
Minimize: & $c_d$ & Coefficient of drag & 1 \\
\midrule
{With respect to:} & $\mathbf{x}$ & FFD control points & 20 \\
& $\alpha$ & Angle of attack & 1 \\
\midrule
{Subject to:} & $c_l = 0.824$ & Lift coefficient & 1 \\
& $c_m = -0.092$ & Pitch moment coefficient & 1 \\
& $V_f \ge V_0$ & Volume constraint & 1 \\
& $\mathbf{t} \ge 0.8\mathbf{t}_0$ & Thickness constraint & 100 \\
\bottomrule
\end{tabularx}
\end{table*}

This work also imposes leading- and trailing-edge constraints in which the corresponding upper and lower FFD control points move with equal magnitude in opposite directions.
These constraints keep the airfoil chord staying on the $x$-axis between (0, 0) and (1, 0). 
In addition, the 20 independent FFD control points are reduced because the paired leading-edge variables and paired trailing-edge variables are not independent.
Each pair is represented by a single independent design variable, resulting in the 18-dimensional FFD representation used throughout this study.
These 18 independent FFD control points, therefore, constitute the airfoil dataset used for training and evaluating GAN models.

The complete dataset contains 992 airfoil designs.
For the present study, both the classical WGAN and Q-WGAN learn a mapping from a three-dimensional latent space to this 18-dimensional FFD design space.
The physical airfoil geometry is recovered by applying the corresponding FFD deformation to the baseline airfoil, producing the final coordinate representation.
Model performance is therefore evaluated in both the 18-dimensional FFD design space and the physical coordinate space.
From the complete dataset, 20 representative airfoils are selected using conditioned Latin hypercube sampling (cLHS)~\cite{minasny2006conditioned}, and the same subset is used for training both models to ensure a fair comparison.

\subsection{Reconstruction and Fitting Performance}

The structures and training details of both classical WGAN and Q-WGAN are shown in Sec.~\ref{sec:method}. 
Table~\ref{tab:fitting_accuracy} summarizes the reconstruction performance of the classical WGAN configuration and the investigated Q-WGAN circuit depths.
The classical WGAN achieves an FFD fitting accuracy of 92.29\%, while the 12-layer Q-WGAN achieves 92.10\%.
The corresponding airfoil-corresponding fitting accuracy are 99.26\% and 99.20\%, respectively.
These results indicate that the Q-WGAN is able to reproduce the target airfoil geometries with reconstruction performance comparable with the classical WGAN despite using a fundamentally different generator architecture.
\begin{table}[!htbp]
\centering
\caption{Fitting accuracy of classical WGAN and Q-WGAN}
\label{tab:fitting_accuracy}
\begin{tabular}{lcc}
\toprule
& FFD control points (\%) & Airfoil coordinates (\%) \\
\midrule
Classical WGAN  & 92.29 & 99.26 \\
Q-WGAN & 92.10 & 99.20 \\
\bottomrule
\end{tabular}
\end{table}

An important observation is the difference between the FFD-control-point and airfoil-coordinate fitting accuracies.
Although the FFD-control-point accuracy is approximately 92\%, the corresponding reconstructed airfoils achieve approximately 99\% coordinate accuracy.
This indicates that deviations in individual FFD design variables do not necessarily produce equally large deviations in the resulting physical airfoil geometry.

\begin{figure}[!h]
\centering
\includegraphics[width=\linewidth]{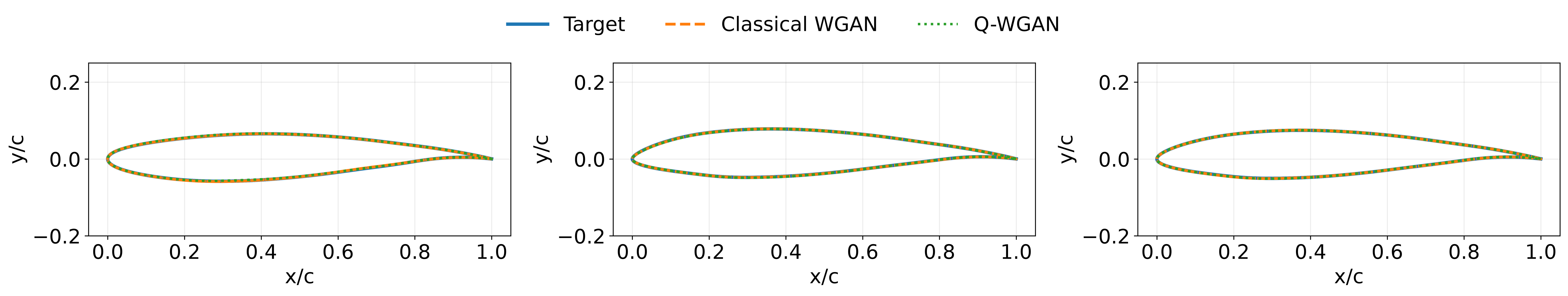}
\caption{Reconstructed airfoils using classical WGAN and Q-WGAN.} 
\label{fig:comp}
\end{figure}

Representative reconstructed airfoils are shown in (Fig.~\ref{fig:comp}).
The reconstructed airfoils by classical WGAN and Q-WGAN are compared against arbitrary reference airfoils. 
Both models closely reproduce the overall airfoil profile, with only minor differences visible in the reconstructed geometry.

\subsection{Distribution Fidelity}

Besides the ability to reconstruct reference airfoils, the trained generative models are evaluated based on distribution metrics, namely, JSD, FD, Wasserstein distance, and KL divergence.
Note that lower values of these metrics indicate closer agreement between the learned and reference distributions.

Table~\ref{tab:dist} shows that Q-WGAN achieves a lower mean JSD of 0.496 than that of the classical WGAN.
Similarly, Q-WGAN obtains $1.67\times10^{-5}$, which is 47.2\% lower than the counterpart of the classical WGAN.
Additionally, the mean Wasserstein distance of Q-WGAN is also lower than that of classical Q-WGAN, presenting a 52.6\% reduction.
These three metrics unanimously and consistently indicate that Q-WGAN-generated data distribution more closely matches the baseline distribution than the classical WGAN.
\begin{table}[!htbp]
\centering
\caption{Distribution metrics comparison for classical WGAN and Q-WGAN}
\label{tab:dist}
\begin{tabular}{lccc}
\toprule
Model & JSD & FD & Wasserstein \\
\midrule
Classical WGAN  & 0.505 & $3.16\times10^{-5}$ & $1.23\times10^{-3}$ \\
Quantum WGAN & 0.496 & $1.67\times10^{-5}$ & $5.09\times10^{-4}$ \\
\bottomrule
\end{tabular}
\end{table}

Table~\ref{tab:kl} presents that the KL-divergence results provide a more mixed comparison because KL divergence is directional.
For the baseline-to-generated direction, the mean KL divergence of the Q-WGAN model is higher than the classical WGAN.
At the individual-variable level, each model achieves the lower KL divergence for 9 of the 18 FFD variables.
Therefore, despite the lower KL divergence by the classical WGAN, neither model consistently outperforms the other across all individual FFD variables.
\begin{table}[!htbp]
\centering
\caption{KL divergence comparison for baseline-to-generated direction}
\label{tab:kl}
\begin{tabular}{lcc}
\toprule
& classical WGAN & Q-WGAN \\
\midrule
Reference $\to$ Generated  & 1.35 & 4.74 \\
Lower KL divergence & 9/18 & 9/18 \\ 
\bottomrule
\end{tabular}
\end{table}

Table~\ref{tab:klinv} exhibits that for the generated-to-baseline direction, the Q-WGAN achieves a slightly lower mean KL divergence compared with the classical WGAN.
In this setup, the Q-WGAN produces lower KL divergence values for 11 of the 18 FFD variables, while the classical WGAN produces the lower value for 7 variables.
Therefore, the Q-WGAN and classical WGAN performance are still comparable with each other. 
\begin{table}[!htbp]
\centering
\caption{KL divergence comparison for generated-to-baseline direction}
\label{tab:klinv}
\begin{tabular}{lcc}
\toprule
& classical WGAN & Q-WGAN \\
\midrule
Generated $\to$ Reference & 2.31 & 2.27 \\
Lower KL divergence & 7/18 & 11/18 \\
\bottomrule
\end{tabular}
\end{table}

The difference between the two KL directions is consistent with the asymmetric nature of KL divergence.
The baseline-to-generated form penalizes regions of the reference distribution that receive insufficient probability coverage from the generated distribution, whereas the reverse direction is more sensitive to generated probability appearing in regions with limited baseline support.
Because the reference set contains only 20 cLHS-selected airfoils, histogram-based KL estimates are also sensitive to bin occupancy and should therefore be interpreted together with the other distribution metrics rather than as an isolated measure of generator quality.

\begin{figure}[H]
\centering
\includegraphics[width=\linewidth]{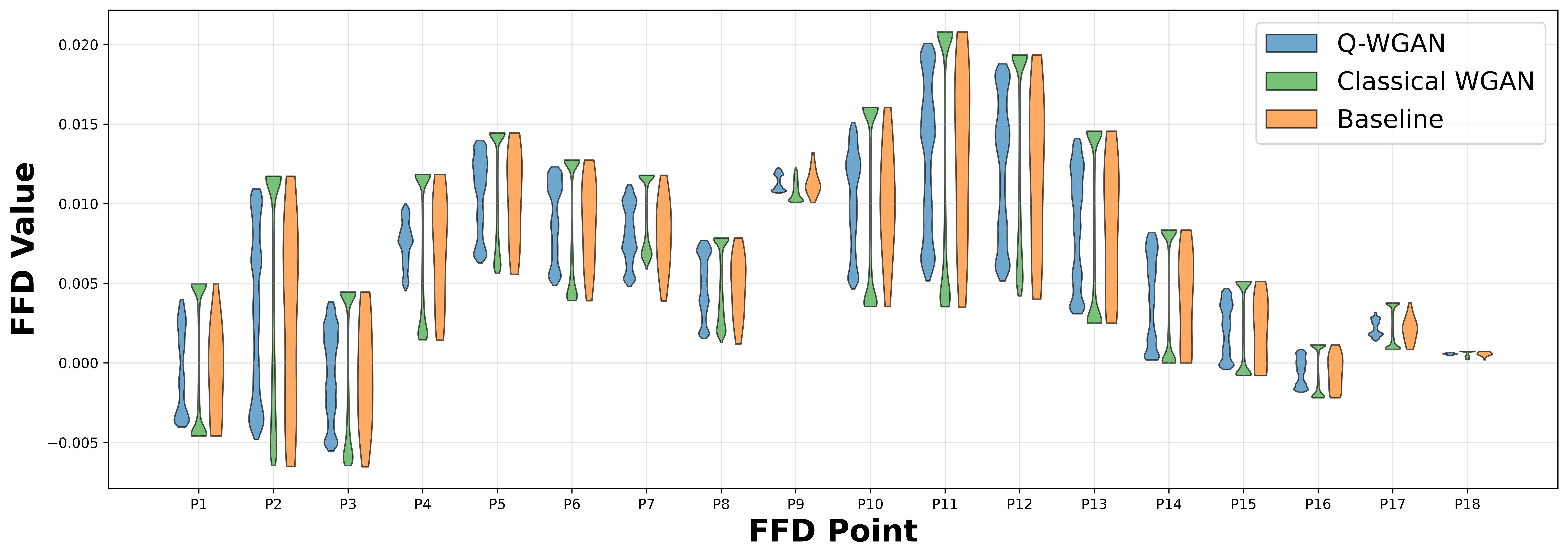}
\caption{Distribution showcase for classical WGAN and Q-WGAN.} 
\label{fig:violin}
\end{figure}

The individual distributions of the 18 FFD variables are shown in Fig.~\ref{fig:violin}.
The classical WGAN exhibits stronger concentration near the boundaries of several FFD-variable ranges, whereas the Q-WGAN generally produces a broader distribution of samples within the interior of the baseline range.
This behavior is consistent with the lower JSD, FD, and Wasserstein distance of Q-WGAN, observed in the quantitative comparison.
However, the individual plots do not fully represent the multivariate relationships among the 18 FFD variables; the joint structure of the generated design spaces is therefore examined separately using t-SNE in the following subsection.

\subsection{Joint Distribution Analysis Using t-SNE}

\begin{figure}[htbp]
\centering
\subfigure[2D t-SNE]{%
    \includegraphics[width=0.48\linewidth]{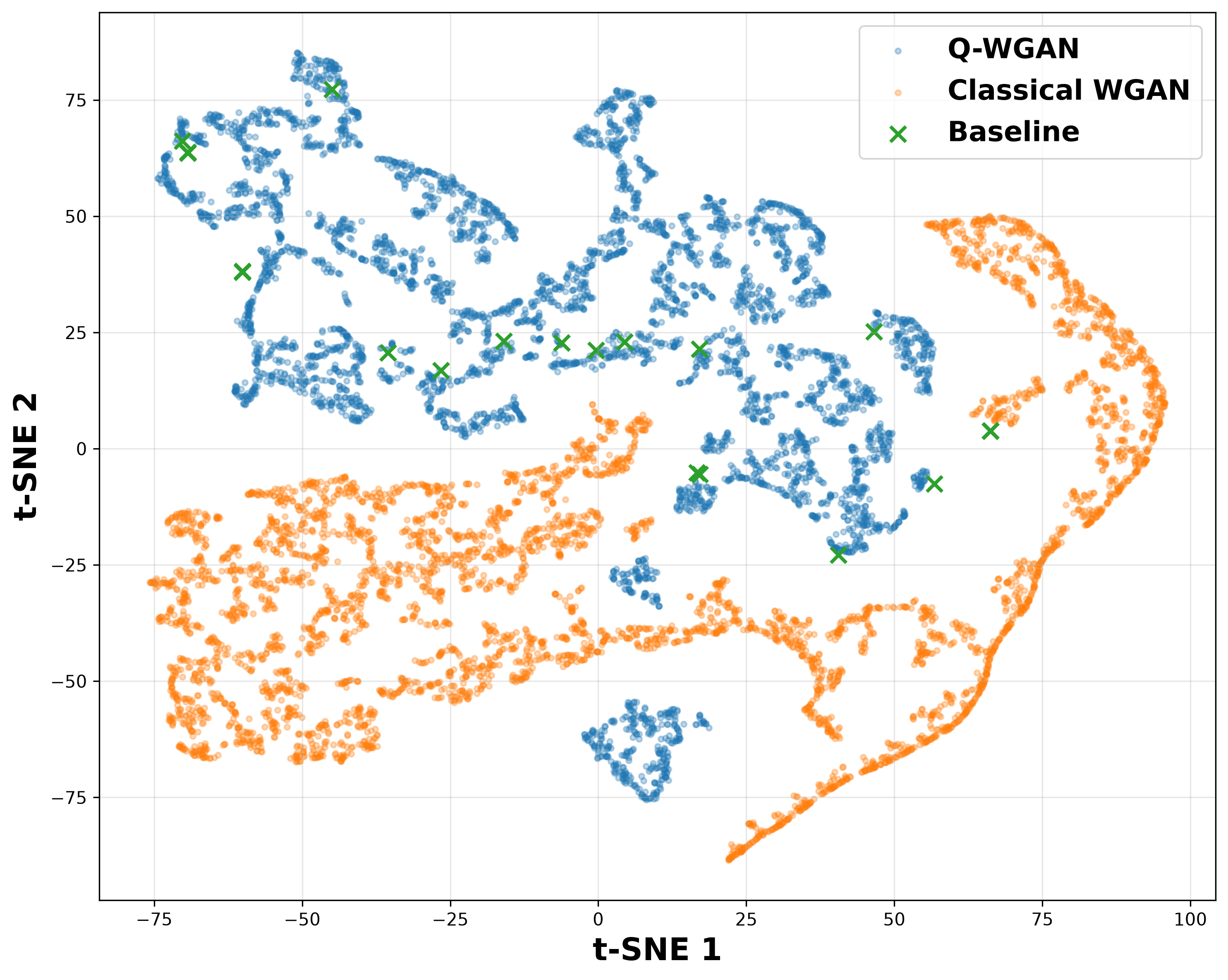}%
    \label{fig:tsne-2d}%
}
\hfill
\subfigure[3D t-SNE]{%
    \includegraphics[width=0.48\linewidth]{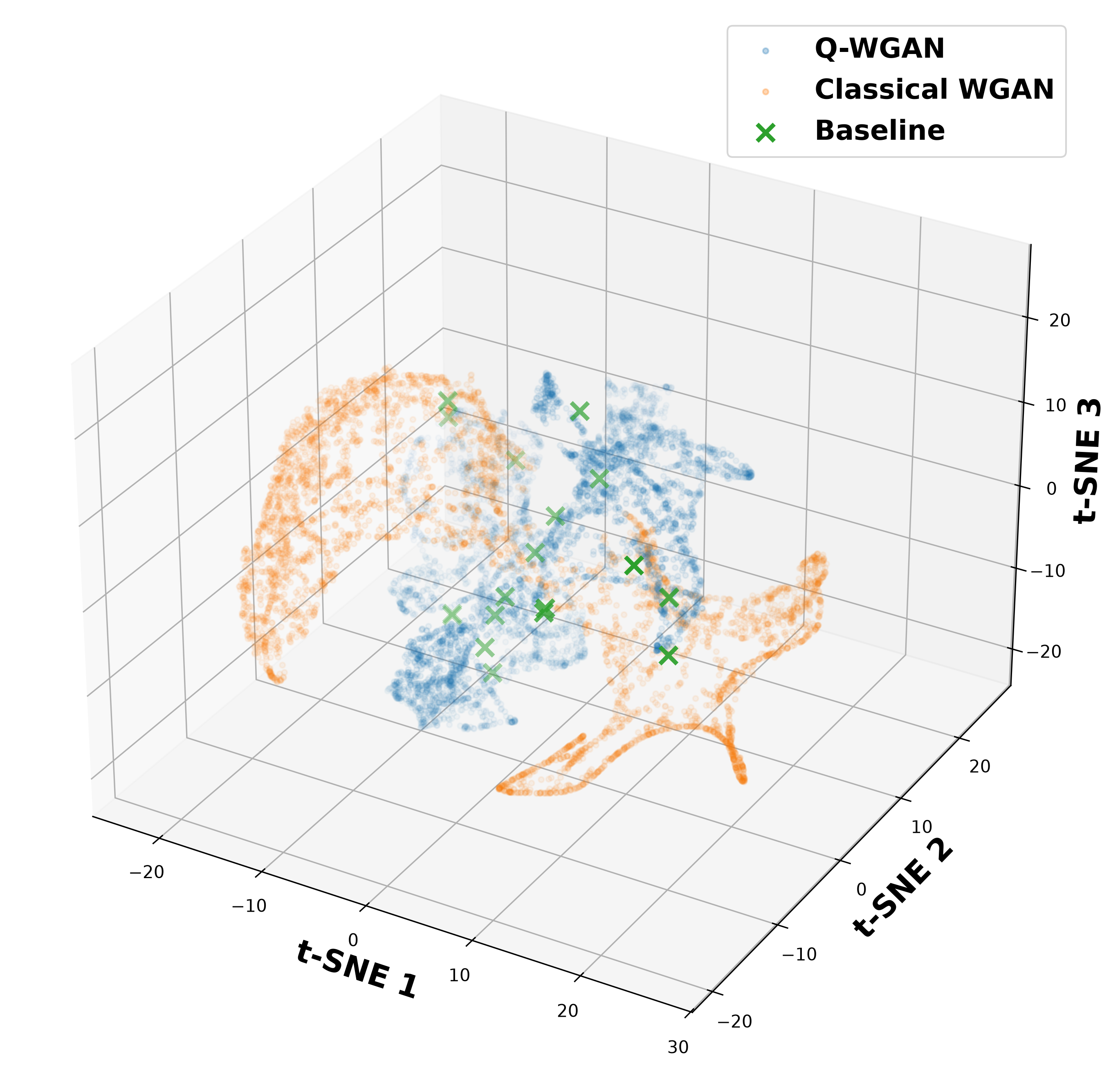}%
    \label{fig:tsne-3d}%
}
\caption{t-SNE visualization of the airfoil parameterization in 2D and 3D spaces.}
\label{fig:tsne}
\end{figure}

To examine the joint structure of the learned airfoil design spaces, t-SNE is applied to the FFD representation, classical WGAN generated, and Q-WGAN generated samples.
Specifically, the FFD variables are first standardized using statistics obtained from the baseline dataset, and the same transformation is applied to the generated samples.
The three datasets are then embedded together into common t-SNE spaces for qualitative comparison of their local neighborhood structures (Fig.~\ref{fig:tsne}).

Figure ~\ref{fig:tsne-2d} shows that the baseline airfoils are distributed across regions occupied by both models.
Several baseline points appear embedded within or close to neighborhoods populated by Q-WGAN samples, while other baseline points lie near classical WGAN-generated regions.
This indicates that neither generative model reproduces the complete baseline structure identically.
However, the Q-WGAN appears to retain stronger local overlap with several of the baseline neighborhoods, whereas the classical WGAN shows greater dispersion into regions that are less densely populated by the baseline samples.

The resulting 3D embedding is shown in Fig. ~\ref{fig:tsne-3d}.
The classical WGAN and Q-WGAN samples occupy noticeably different regions and form distinct structures in the embedded design space, indicating that the two generators learn different representations of the original FFD distribution.
The Q-WGAN samples are concentrated more strongly through the central region of the embedding, while the classical WGAN samples form broader and more separated branches extending across the t-SNE space.

These observations are consistent with the quantitative distribution metrics presented in the previous subsection.
The Q-WGAN achieves lower JSD, FD, and Wasserstein distance than the classical WGAN, suggesting closer overall agreement with the baseline distribution under the selected configurations.
At the same time, the t-SNE visualization reveals that the two models differ also in the structure of the design spaces they generate.

\section{Conclusion}
\label{sec:conclusion}

This study developed a hybrid quantum Wasserstein generative adversarial networks (Q-WGAN), consisting of quantum generator networks and classical discriminator networks, for airfoil parameterization.
Both models were trained using only 20 optimal airfoils and mapped a three-dimensional latent space to the free-form deformation (FFD) control-point space.
The trained Q-WGAN was then evaluated using reconstruction accuracy, distribution-based metrics, and t-distributed stochastic neighbor embedding visualization, and compared against classifical Wasserstein generative adversarial networks (WGAN).

The results showed that both classical WGAN and Q-WGAN models were able to reconstruct reference airfoil shapes with high accuracy.
The Q-WGAN achieved almost the same fitting accuracies on FFD control points and airfoil coordinates. 
Moreover, in the distribution metric comparisons, the Q-WGAN obtained lower Jensen–Shannon divergence, Fréchet distance, and Wasserstein distance, representing a closer match with the baseline distribution. 
The Kullback-Leibler divergence values were mixed depending on the direction of comparison, showing a comparable performance between classical WGAN and Q-WGAN. 
The t-distributed stochastic neighbor embedding analysis also showed that the classical WGAN and Q-WGAN learned different structures in the design space, with the Q-WGAN showing stronger overlap with the baseline distribution.

Overall, the results indicated that the Q-WGAN can provide reconstruction performance comparable to the classical WGAN while using a much smaller generator architecture.
At the same time, the distribution analyses suggested that the Q-WGAN captured distribution characteristics of the baseline design space using a highly limited training dataset.
These results would support further applications of Q-WGAN for aerodynamic shape optimization.

Future work will focus on evaluating the method using larger and more diverse airfoil datasets, studying the effect of training-set size, and extending the framework toward aerodynamic optimization tasks with additional physical and performance constraints.
\bibliography{sample}

@article{li2022machine,
  title={Machine learning in aerodynamic shape optimization},
  author={Li, Jichao and Du, Xiaosong and Martins, Joaquim RRA},
  journal={Progress in Aerospace Sciences},
  volume={134},
  pages={100849},
  year={2022},
  publisher={Elsevier}
}

@article{yasong2018global,
  title={Global aerodynamic design optimization based on data dimensionality reduction},
  author={Yasong, QIU and Junqiang, BAI and Nan, LIU and Chen, WANG},
  journal={Chinese Journal of Aeronautics},
  volume={31},
  number={4},
  pages={643--659},
  year={2018},
  publisher={Elsevier}
}

@article{chen2020airfoil,
  title={Airfoil design parameterization and optimization using b{\'e}zier generative adversarial networks},
  author={Chen, Wei and Chiu, Kevin and Fuge, Mark D},
  journal={AIAA journal},
  volume={58},
  number={11},
  pages={4723--4735},
  year={2020},
  publisher={American Institute of Aeronautics and Astronautics}
}

@inproceedings{chen2019aerodynamic,
  title={Aerodynamic design optimization and shape exploration using generative adversarial networks},
  author={Chen, Wei and Chiu, Kevin and Fuge, Mark},
  booktitle={AIAA Scitech 2019 forum},
  pages={2351},
  year={2019}
}

@inproceedings{chen2021deep,
  title={Deep generative model for efficient 3D airfoil parameterization and generation},
  author={Chen, Wei and Ramamurthy, Arun},
  booktitle={AIAA Scitech 2021 Forum},
  pages={1690},
  year={2021}
}

@article{viswanath2011dimension,
  title={Dimension reduction for aerodynamic design optimization},
  author={Viswanath, Asha and Forrester, Alexander IJ and Keane, AJ},
  journal={AIAA journal},
  volume={49},
  number={6},
  pages={1256--1266},
  year={2011}
}

@article{alswaitti2022dimensionality,
  title={Dimensionality reduction, modelling, and optimization of multivariate problems based on machine learning},
  author={Alswaitti, Mohammed and Siddique, Kamran and Jiang, Shulei and Alomoush, Waleed and Alrosan, Ayat},
  journal={Symmetry},
  volume={14},
  number={7},
  pages={1282},
  year={2022},
  publisher={MDPI}
}

@article{jun2020application,
  title={Application of a PCA-DBN-based surrogate model to robust aerodynamic design optimization},
  author={Jun, TAO and Gang, SUN and Liqiang, GUO and Xinyu, WANG},
  journal={Chinese Journal of Aeronautics},
  volume={33},
  number={6},
  pages={1573--1588},
  year={2020},
  publisher={Elsevier}
}

@article{cinquegrana2018investigation,
  title={Investigation of adaptive design variables bounds in dimensionality reduction for aerodynamic shape optimization},
  author={Cinquegrana, Davide and Iuliano, Emiliano},
  journal={Computers \& Fluids},
  volume={174},
  pages={89--109},
  year={2018},
  publisher={Elsevier}
}

@article{tejero2023aerodynamic,
  title={Aerodynamic optimisation of civil aero-engine nacelles by dimensionality reduction and multi-fidelity techniques},
  author={Tejero, Fernando and MacManus, David and Hueso-Rebassa, Josep and Sanchez-Moreno, Francisco and Goulos, Ioannis and Sheaf, Christopher},
  journal={International Journal of Numerical Methods for Heat \& Fluid Flow},
  volume={33},
  number={4},
  pages={1336--1353},
  year={2023},
  publisher={Emerald Publishing Limited}
}

@article{wu2024high,
  title={High-dimensional aerodynamic shape optimization framework using geometric domain decomposition and data-driven support strategy for wing design},
  author={Wu, Xiaojing and Ma, Long and Zuo, Zijun},
  journal={Aerospace Science and Technology},
  volume={149},
  pages={109152},
  year={2024},
  publisher={Elsevier}
}

@article{bo2024data,
  title={Data-driven surrogate model for aerodynamic design using separable shape tensor method},
  author={Bo, PANG and Zhang, Yang and Junlin, LI and Xudong, WANG and Chang, Min and Junqiang, BAI},
  journal={Chinese Journal of Aeronautics},
  volume={37},
  number={9},
  pages={41--58},
  year={2024},
  publisher={Elsevier}
}

@inproceedings{du2020b,
  title={A B-spline-based generative adversarial network model for fast interactive airfoil aerodynamic optimization},
  author={Du, Xiaosong and He, Ping and Martins, Joaquim RRA},
  booktitle={AIAA scitech 2020 forum},
  pages={2128},
  year={2020}
}

@article{wang2021airfoil,
  title={Airfoil GAN: Encoding and synthesizing airfoils for aerodynamic-aware shape optimization},
  author={Wang, Yuyang and Shimada, Kenji and Farimani, Amir Barati},
  journal={arXiv preprint arXiv:2101.04757},
  year={2021}
}

@inproceedings{sisk2026physics,
  title={Physics-Constrained Generative Adversarial Networks for Dimensionality Reduction in Optimization},
  author={Sisk, Sam and Du, Xiaosong},
  booktitle={AIAA SCITECH 2026 Forum},
  pages={2016},
  year={2026}
}

@article{goodfellow2014generative,
  title={Generative adversarial nets},
  author={Goodfellow, Ian J and Pouget-Abadie, Jean and Mirza, Mehdi and Xu, Bing and Warde-Farley, David and Ozair, Sherjil and Courville, Aaron and Bengio, Yoshua},
  journal={Advances in neural information processing systems},
  volume={27},
  year={2014}
}

@inproceedings{wu2019pwgan,
  title={PWGAN: Wasserstein GANs with perceptual loss for mode collapse},
  author={Wu, Xianyu and Shi, Canghong and Li, Xiaojie and He, Jia and Wu, Xi and Lv, Jiancheng and Zhou, Jiliu},
  booktitle={Proceedings of the ACM Turing Celebration Conference-China},
  pages={1--7},
  year={2019}
}

@inproceedings{chen2020improvement,
  title={An improvement based on Wasserstein GAN for alleviating mode collapsing},
  author={Chen, Yingying and Hou, Xinwen},
  booktitle={2020 international joint conference on neural networks (IJCNN)},
  pages={1--7},
  year={2020},
  organization={IEEE}
}

@article{ngo2023survey,
  title={A survey of recent advances in quantum generative adversarial networks},
  author={Ngo, Tuan A and Nguyen, Tuyen and Thang, Truong Cong},
  journal={Electronics},
  volume={12},
  number={4},
  pages={856},
  year={2023},
  publisher={MDPI}
}

@article{chakrabarti2019quantum,
  title={Quantum Wasserstein generative adversarial networks},
  author={Chakrabarti, Shouvanik and Yiming, Huang and Li, Tongyang and Feizi, Soheil and Wu, Xiaodi},
  journal={Advances in Neural Information Processing Systems},
  volume={32},
  year={2019}
}

@inproceedings{arjovsky2017wasserstein,
  title={Wasserstein generative adversarial networks},
  author={Arjovsky, Martin and Chintala, Soumith and Bottou, L{\'e}on},
  booktitle={International conference on machine learning},
  pages={214--223},
  year={2017},
  organization={Pmlr}
}

@article{gulrajani2017improved,
  title={Improved training of wasserstein gans},
  author={Gulrajani, Ishaan and Ahmed, Faruk and Arjovsky, Martin and Dumoulin, Vincent and Courville, Aaron},
  journal={Advances in neural information processing systems},
  volume={30},
  year={2017}
}

@article{yonekura2024airfoil,
  title={Airfoil shape generation and feature extraction using the conditional vae-wgan-gp},
  author={Yonekura, Kazuo and Tomori, Yuki and Suzuki, Katsuyuki},
  journal={AI},
  volume={5},
  number={4},
  pages={2092--2103},
  year={2024},
  publisher={MDPI}
}

@article{lloyd2018quantum,
  title={Quantum generative adversarial learning},
  author={Lloyd, Seth and Weedbrook, Christian},
  journal={Physical review letters},
  volume={121},
  number={4},
  pages={040502},
  year={2018},
  publisher={APS}
}

@article{situ2020quantum,
  title={Quantum generative adversarial network for generating discrete distribution},
  author={Situ, Haozhen and He, Zhimin and Wang, Yuyi and Li, Lvzhou and Zheng, Shenggen},
  journal={Information Sciences},
  volume={538},
  pages={193--208},
  year={2020},
  publisher={Elsevier}
}

@article{zoufal2019quantum,
  title={Quantum generative adversarial networks for learning and loading random distributions},
  author={Zoufal, Christa and Lucchi, Aur{\'e}lien and Woerner, Stefan},
  journal={npj Quantum Information},
  volume={5},
  number={1},
  pages={103},
  year={2019},
  publisher={Nature Publishing Group UK, London}
}

@article{huang2021experimental,
  title={Experimental quantum generative adversarial networks for image generation},
  author={Huang, He-Liang and Du, Yuxuan and Gong, Ming and Zhao, Youwei and Wu, Yulin and Wang, Chaoyue and Li, Shaowei and Liang, Futian and Lin, Jin and Xu, Yu and others},
  journal={Physical Review Applied},
  volume={16},
  number={2},
  pages={024051},
  year={2021},
  publisher={APS}
}

@inproceedings{kushwaha2020study,
  title={Study of prevention of mode collapse in generative adversarial network (GAN)},
  author={Kushwaha, Vandana and Nandi, GC and others},
  booktitle={2020 IEEE 4th Conference on Information \& Communication Technology (CICT)},
  pages={1--6},
  year={2020},
  organization={IEEE}
}

@article{storn1997differential,
  title={Differential evolution--a simple and efficient heuristic for global optimization over continuous spaces},
  author={Storn, Rainer and Price, Kenneth},
  journal={Journal of global optimization},
  volume={11},
  number={4},
  pages={341--359},
  year={1997},
  publisher={Springer}
}

@article{lin1991divergence,
  title={Divergence measures based on the Shannon entropy},
  author={Lin, Jianhua},
  journal={IEEE Transactions on Information theory},
  volume={37},
  number={1},
  pages={145--151},
  year={1991},
  publisher={IEEE}
}

@article{heusel2017gans,
  title={Gans trained by a two time-scale update rule converge to a local nash equilibrium},
  author={Heusel, Martin and Ramsauer, Hubert and Unterthiner, Thomas and Nessler, Bernhard and Hochreiter, Sepp},
  journal={Advances in neural information processing systems},
  volume={30},
  year={2017}
}

@article{kullback1951information,
  title={On information and sufficiency},
  author={Kullback, Solomon and Leibler, Richard A},
  journal={The annals of mathematical statistics},
  volume={22},
  number={1},
  pages={79--86},
  year={1951},
  publisher={JSTOR}
}

@article{van2008visualizing,
  title={Visualizing data using t-SNE.},
  author={Van der Maaten, Laurens and Hinton, Geoffrey},
  journal={Journal of machine learning research},
  volume={9},
  number={11},
  year={2008}
}

@article{martins2022aerodynamic,
  title={Aerodynamic design optimization: Challenges and perspectives},
  author={Martins, Joaquim RRA},
  journal={Computers \& Fluids},
  volume={239},
  pages={105391},
  year={2022},
  publisher={Elsevier}
}

@article{kingma2014adam,
  title={Adam: A method for stochastic optimization},
  author={Kingma, Diederik P and Ba, Jimmy},
  journal={arXiv preprint arXiv:1412.6980},
  year={2014}
}

@article{hajdik2023pygeo,
  title={pyGeo: A geometry package for multidisciplinary design optimization},
  author={Hajdik, Hannah M and Yildirim, Anil and Wu, Ella and Brelje, Benjamin J and Seraj, Sabet and Mangano, Marco and Anibal, Joshua L and Jonsson, Eirikur and Adler, Eytan J and Mader, Charles A and others},
  journal={Journal of Open Source Software},
  volume={8},
  number={87},
  pages={5319},
  year={2023}
}

@article{mader2020adflow,
  title={ADflow: An open-source computational fluid dynamics solver for aerodynamic and multidisciplinary optimization},
  author={Mader, Charles A and Kenway, Gaetan KW and Yildirim, Anil and Martins, Joaquim RRA},
  journal={Journal of Aerospace Information Systems},
  volume={17},
  number={9},
  pages={508--527},
  year={2020},
  publisher={American Institute of Aeronautics and Astronautics}
}

@article{minasny2006conditioned,
  title={A conditioned Latin hypercube method for sampling in the presence of ancillary information},
  author={Minasny, Budiman and McBratney, Alex B},
  journal={Computers \& geosciences},
  volume={32},
  number={9},
  pages={1378--1388},
  year={2006},
  publisher={Elsevier}
}

\end{document}